\documentclass[conference]{IEEEtran}
\usepackage{cite}
\usepackage[hidelinks]{hyperref}

\ifCLASSINFOpdf
  \usepackage[pdftex]{graphicx}
\else
  \usepackage[dvips]{graphicx}
\fi
\usepackage[cmex10]{amsmath}
\usepackage[algoruled]{algorithm2e}
\usepackage{url}

\begin{document}
%
\title{A 3D Reactive Navigation Algorithm for Mobile Robots by Using Tentacle-Based Sampling}

\author{\IEEEauthorblockN{Ne\c{s}et \"{U}nver Akmandor}
\IEEEauthorblockA{Department of Electrical and Computer Engineering\\
Northeastern University, Boston, MA, 02115, USA\\
Email: akmandor.n@northeastern.edu}
\and
\IEEEauthorblockN{Ta\c{s}k{\i}n Pad{\i}r}
\IEEEauthorblockA{Institute for Experiential Robotics\\
Northeastern University, Boston, MA, USA\\
Email: t.padir@northeastern.edu}}


%


\maketitle

\begin{abstract}
This paper introduces a reactive navigation framework for mobile robots in 3-dimensional (3D) space. The proposed approach does not rely on the global map information and achieves fast navigation by employing a tentacle-based sampling and their heuristic evaluations on-the-fly. This reactive nature of the approach comes from the prior arrangement of navigation points on tentacles (parametric contours) to sample the navigation space. These tentacles are evaluated at each time-step, based on heuristic features such as closeness to the goal, previous tentacle preferences and nearby obstacles in a robot-centered 3D grid. Then, the navigable sampling point on the selected tentacle is passed to a controller for the motion execution. The proposed framework does not only extend its 2D tentacle-based counterparts into 3D, but also introduces offline and online parameters, whose tuning provides versatility and adaptability of the algorithm to work in unknown environments. To demonstrate the superior performance of the proposed algorithm over a state-of-art method, the statistical results from physics-based simulations on various maps are presented. The video of the work is available at \url{https://youtu.be/rrF7wHCz-0M}.
\end{abstract}

\IEEEpeerreviewmaketitle

\section{Introduction}
Towards realizing fully autonomous robots, motion and path planning remains to be an active and still challenging research direction in robotics. Especially because of their mobility and flexibility, the real-world applications with unmanned aerial vehicles (UAVs) have become the focus of many academic \cite{lin2018autonomous,oleynikova2016continuous} or industrial projects \cite{beul2018fast,escobar2018r}. Based on their use cases, these applications involve highly challenging tasks, such as mapping \cite{oleynikova2019open} and safe path planning \cite{gao2018online,lin2018autonomous,usenko2017real,oleynikova2016continuous}, which commonly require considerable amount of memory to store the data and computational power to process them. In order to meet these requirements and focusing on the autonomous navigation problem in unknown environments, researchers \cite{campos2019autonomous,mohta2018fast} develop algorithms that are capable of both working on onboard systems and performing online data processing. ~\\

\subsection{Contribution}
In a real-world scenario, one of the biggest challenges in autonomous navigation problem is the lack of the prior knowledge of a global map. Even the map is available, the dynamic nature of the environment makes this prior info impractical to use as a reliable source. To solve this problem, we propose a reactive path planning framework by extending tentacle-based navigation for 3D environment. Without using any prior global map and planning the entire path, at each iteration the robot's next pose is determined by the evaluation of the pre-calculated sampling points. To the best of our knowledge, this is the first use of tentacle-based sampling within a reactive navigation framework for 3D environments. We divide the ego-centered volume around the robot into voxels to provide direct mapping of the occupancy data into a 3D grid. As our second contribution, we introduce offline and online parameters to enhance navigation performance in unknown environments. The methodologies of tuning these parameters are discussed throughout the paper. Third, we provide the implementation details including computational complexity analysis to enable the reproducibility of the algorithm. Last, we compared our algorithm with the state-of-art method using benchmark map datasets. Overall, our proposed reactive algorithm outperforms two configurations of the other method in terms of success rate and navigation duration. The open-source implementation of the algorithm and the benchmarks can be found at \url{https://github.com/RIVeR-Lab/tentabot}. ~\\

\subsection{Related Work}
Authors in \cite{usenko2017real} keep the local occupancy information around the robot by a 3D circular buffer and adjust the local trajectory represented by a B-spline. Despite having the possibility of getting stuck at the local minima, the parameters of the B-spline is calculated by optimizing a cost function which pulls the robot towards goal and drive away from obstacles while keeping the robot's motion stable. Lin et al. \cite{lin2018autonomous} and Gao et al. \cite{gao2018online} require high computation power due to their image processing and optimization steps. Both framework estimate the 3D local map using the data from camera and inertial measurement unit. Based on the map, the work in \cite{lin2018autonomous} generates the local path by a sampling-based algorithm, RRG \cite{karaman2011sampling}. Differently, Gao et al. \cite{gao2018online} calculate Euclidean Signed Distance Field and applies fast marching method to obtain the path. Initializing with a given path, the non-linear optimization solver ensures the smoothness and dynamical feasibility of the final trajectory for each method. In \cite{mohta2018fast}, Mohta et al. propose a trajectory planner in GPS-denied and cluttered environments, providing detailed aspects on both hardware and software. Similar to the aforementioned algorithms, they also combine a sampling-based method, A* \cite{hart1968formal}, with an optimization process to generate the robot trajectory. To avoid local minima during trajectory calculation, they propose a combined map structure that keeps the local occupancy information in 3D while the global one is in 2D. However, even though the global map is planar, the size of the map and discrete nature of the A* algorithm limit their framework for the real-world scenarios. Being one of the most recent works in the autonomous navigation context, Oleynikova et al. \cite{oleynikova2019open} propose a framework for mapping, planning and trajectory generation. Having vision based sensing, they compute the Truncated Signed Distance Field to project the environment around the robot into a map which represents the collision costs. For the path planning, they first generate a deterministic graph in the free-space of their map and then find the path using A*. In the last step, the trajectory of the robot is calculated by the optimization considering the trade-off between reaching to the goal and exploration. In another recent Micro Aerial Vehicle (MAV) framework \cite{campos2019autonomous}, the local occupancy information is represented by linear octree structure. Following that, the motion of the robot is planned by RRT-Connect \cite{karaman2011sampling}. The trajectory generation, which includes an offline stage of LQR virtual control design and Lyapunov analysis, guarantees that the dynamic constraints are satisfied. \\

The idea of reactive navigation has emerged to traverse dynamic environments where agent does not have a prior global map but only the local sensor information. Escobar et al. \cite{escobar2018r} and Beul et al. \cite{beul2018fast} use visual perception and reactive control algorithms to avoid obstacles and achieve fast navigation towards the goal with a Unmanned Aerial Vehicle (UAV) system. Their approaches differ from each other such that Escobar et al. use potential fields to reach the goal, while Beul et al. plan a path of poses using the integration of A* and Ramer-Douglas Peucker algorithms. For a 2D action space, the reactive navigation algorithm \cite{von2008driving} of the 2007 European Land Robot Trial winner and DARPA Urban Challenge finalist team enables fast navigation towards to a goal while avoiding obstacles in highly cluttered environments. In their paper, Von Hundelshausen et. al. refer pre-calculated trajectories as tentacles which are formed with respect to vehicle's coordinate frame. Additionally, they present a methodology to use these tentacles as perceptual primitives to map occupancy grid information into a tentacle (trajectory) selection. Later, they extend their previous work by accumulating LIDAR data into multi-layered occupancy grid in \cite{himmelsbach2009team} and updating their circular tentacle form to clothoid considering steering angle in \cite{himmelsbach2011autonomous}. Integrating their robot's kinematics into circular tentacle calculation, Cherubini et al. \cite{cherubini2012new} use visual data for navigation while avoiding static obstacles. Then they perform dynamic obstacle avoidance in their following paper \cite{cherubini2014autonomous}. The work in \cite{alia2015local} forms clothoid version of tentacles and the selected tentacle is performed by their vehicle using a lateral controller based on Immersion and Invariance principle. Similarly, forming clothoid trajectories, study in \cite{mouhagir2016markov} decides best tentacle at each step by Markov Decision Process and in \cite{mouhagir2017trajectory} they map occupancy information into an evidential grid structure which enables to represent sensor based uncertainties. Instead of a path planner, Zhang et al. \cite{zhang2017formation} use tentacle concept to ensure multiple UAV flight formation and reactive obstacle avoidance. Most recently, Khelloufi et al. \cite{khelloufi2017tentacle} propose a tentacle-based obstacle avoidance scheme for omni-directional mobile robots which can visually track a target while navigating. ~\\

\section{3D Reactive Navigation Framework}

\subsection{Context} \label{sec:context} 

Our navigation framework is defined in 3D workspace which is assumed to consist of either free or occupied subspaces in a fixed Cartesian coordinate frame $W$. The occupied space contains both static and dynamic objects including our robot. In order to locate these objects and update their recent positions, $p^W_{(x,y,z)}$, let us also define the local robot frame $R$ and the sensor frame $S$ with respect to $W$. ~\\
    	
The main objective of our algorithm is to find a navigable path from a start position $p^{start}$ to a goal position $p^{goal}$, meanwhile satisfying multiple objectives such as; closest proximity to the goal, collision-free path and minimum navigation time.

\subsection{Robot-Centered 3D Grid} \label{sec:3dgrid} 
    	
\begin{figure}[!t]
\centering
\includegraphics[width=3.5in]{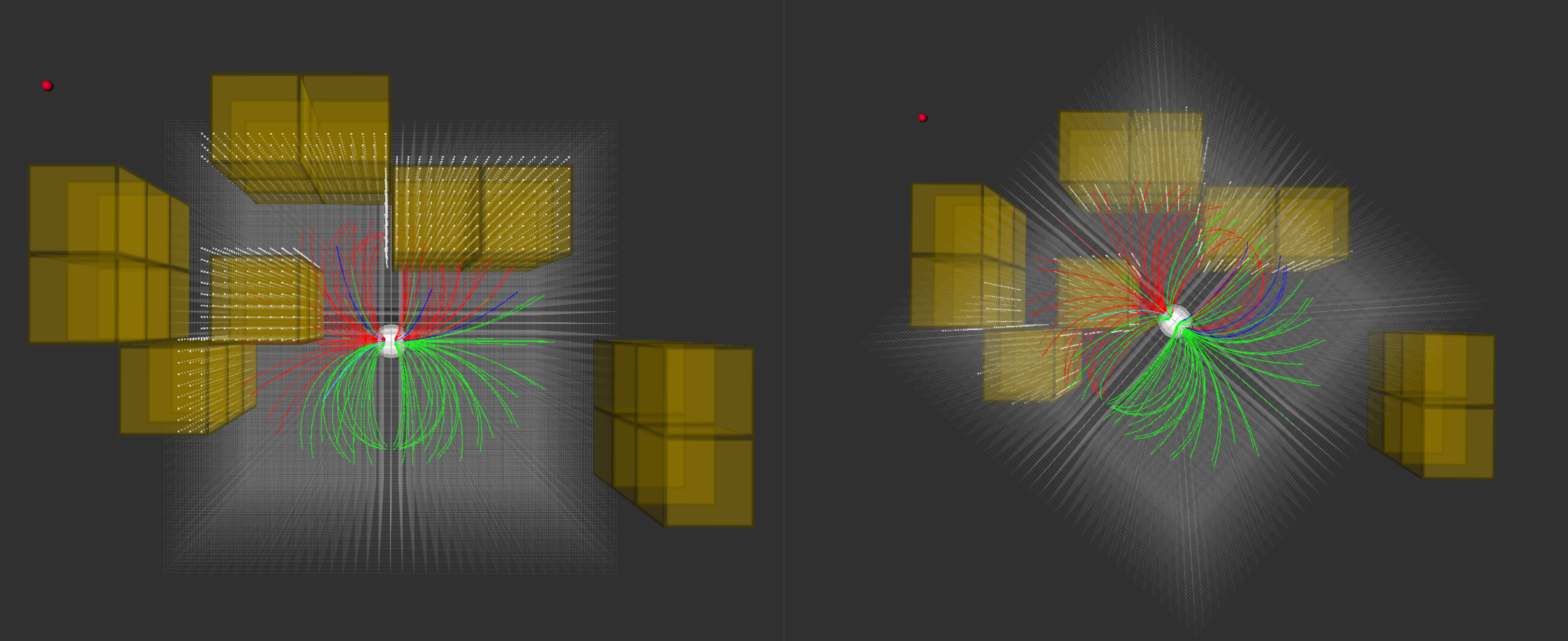}
\caption{The robot-centered grid $G$ (shaded grey region) is formed by $N^v$ voxels with dimension $d^v$. In each time step, local occupancy info around the robot is mapped into $G$. While navigating towards the goal (red sphere), only obstacles (yellow cubes) inside $G$ is considered. Tentacles are formed by the group of pre-calculated sampling points that are fixed to robot's coordinate frame. The occupancy around the robot determines whether the tentacle is navigable (green), non-navigable (red) or temporarily navigable (blue).}
\label{fig:robot_grid}
\end{figure}

Enhancing the 2D approach from \cite{von2008driving}, 3D grid, $G$, is formed around the robot by aligning both coordinate frames as shown in Fig. \ref{fig:robot_grid}. The robot-centered grid is composed of $N^{v}$ cubic voxels. The number of voxels, $n^{v}_{\{x,y,z\}}$, for each axes is determined by $N^{v} = n^{v}_z n^{v}_y n^{v}_z$. The width, length and height ${\{w,l,h\}}^G$ of the grid is calculated as ${\{w,l,h\}}^G = d^{v} n^{v}_{\{x,y,z\}}$ by the given voxel dimension, $d^{v}$. ~\\

As an input to our framework, the point cloud data, $D$, is assumed to be received at a specified frequency $f^S$. This data could be obtained by any sensor that measures spatial occupancy information around the robot and it is assumed in the form of $D = \{[p^S_m(x,y,z), \rho_m] \quad | \quad m=1,...,N^D\}$. Here, $p^S_m(x,y,z)$ is the coordinate of an occupied point with respect to sensor frame $S$ and $\rho_m$ is the probabilistic belief value that sensor supplies. Each received occupancy point $m$ is mapped into its respective voxel which keeps the average belief value as $\rho_{avg}$. Although our proposed framework does not keep a global map, we store some history of the point cloud data to compensate the lack of the occupancy information in the close range of the robot. Taking into account the minimum range of the sensor starts from some threshold, the occupancy history plays a crucial role to avoid obstacles, especially when the robot changes its orientation rapidly.  ~\\

Since our proposed algorithm is designed to navigate in 3D space unlike in \cite{von2008driving}, we need to keep point cloud information without any planar mapping. On the other hand, considering memory efficiency, our algorithm also keeps the point cloud data, in a linear array format. The mapping, $M: SE(3) \rightarrow SE(1)$, from Cartesian coordinates to linearized index is given in the Eq. (\ref{eqn:lin}) where $\{x,y,z\}$ are given with respect to $R$ in light of the coordinate frame transformation from $S$. From the programming perspective, this data could also be stored by a memory efficient tree structure Octomap \cite{hornung2013octomap}. However, since run-time of a search in this tree structure has $O(nlog(n))$ compared to constant $O(1)$ time in linear array, we prefer faster call over memory efficiency in our implementation.
\begin{subequations}
\label{eqn:lin}
\begin{align} 
A(o_i) &= \rho_{avg}\text{,} \quad \text{where}
\end{align}
\begin{align}
o_i &= o_{i_x} + o_{i_y} n^{v}_x + o_{i_z} n^{v}_x n^{v}_y \\
o_{i_{\{x,y,z\}}} &= \frac{n^{v}_{\{x,y,z\}}}{2} + floor(\frac{{\{x,y,z\}}}{d^v}). 
\end{align}
\end{subequations}

\subsection{Tentacles} \label{sec:tentacle}
    	
Tentacles are pre-calculated paths that are fixed to robot's coordinate frame starting from the volumetric center of the 3D grid. Assuming the constant lateral and angular velocity, \cite{von2008driving} generates these tentacles as circular arcs since drivable paths of their "bicycle modeled" ground vehicle are circular. When omni-directional robots are considered, linear paths can be considered as the common ground since they sample the space more uniformly than its counterparts and they are simpler in terms of computation. Although it is not strictly necessary, generating these tentacles by considering the dynamical structure of the robotic platforms tends to improve the performance of the navigation algorithm. The generated tentacles do not necessarily match with the feasible path solutions, because they are also used to sense the environment. Hence, instead of kinodynamically sampling the tentacles, in our framework the feasibility of the selected path is left to be validated by the motion execution block. ~\\

Each tentacle is formed by the sampling points, $p^{R}_{(x,y,z)}$, which are initiated on the $xy$-plane with respect to robot's coordinate frame, $R$. Each tentacle has $l^t$ length and is formed by $n^{s}$ sampling points. The angular coverage, $\varphi$, of total tentacles along the yaw ($z$-axis) is sampled by $n_{\varphi}$ number of tentacles. Then these planar tentacles are extended to 3D by either rotating around pitch ($x$-axis) or roll ($y$-axis). Hence, the respective $\theta$ or $\psi$ angles are sampled by either $n_{\theta}$ or $n_{\psi}$ number of tentacles. For each tentacle $j$, the position of each sampling point $p^{R}_k{(x,y,z)}$ are stored in the set $T_j$. Hence, the total set, $Q$, of $N^t$ tentacles contains $N^s$ sampling points as shown in Eq. (\ref{eqn:tentacle_set}), where $N^t = \{n_{\varphi} n_{\phi}| \phi \in \{\theta, \psi\}\}$ and $N^s = N^t n^s$.
\begin{subequations}
    \begin{align}
        Q &= \{T_j \quad | \quad j = 1,...,N^t\} \\
        \quad T_j &= \{p^{R}_k(x,y,z) \quad | \quad k = 1,...,n^s\}.
    \end{align}
\label{eqn:tentacle_set}
\end{subequations}

\subsection{Support and Priority Voxels} \label{sec:support}

For each tentacle $j$, the set of voxels are determined based on the distance between the sampling points on the tentacle and voxel positions in $G$. The voxel structure consists of four variables where $v=(o, \beta, s, c)$. These variables are adjusted prior to the navigation to enable fast computation of the heuristic values. The first variable, $o$, is the index of the of the corresponding voxel position in the linearized array, $A$. The second variable, $\beta$, keeps the occupancy weight based on the shortest distance between the voxel and the $j^{th}$ tentacle. The third variable, $s$ holds the index of the closest sampling point on the $j^{th}$ tentacle to the voxel. Last variable, $c$ indicates the class type of the voxel which can be either Priority ($c = 1$) or Support ($c = 0$). ~\\
    	
\begin{figure}[!ht]
\centering
\includegraphics[width=3.5in]{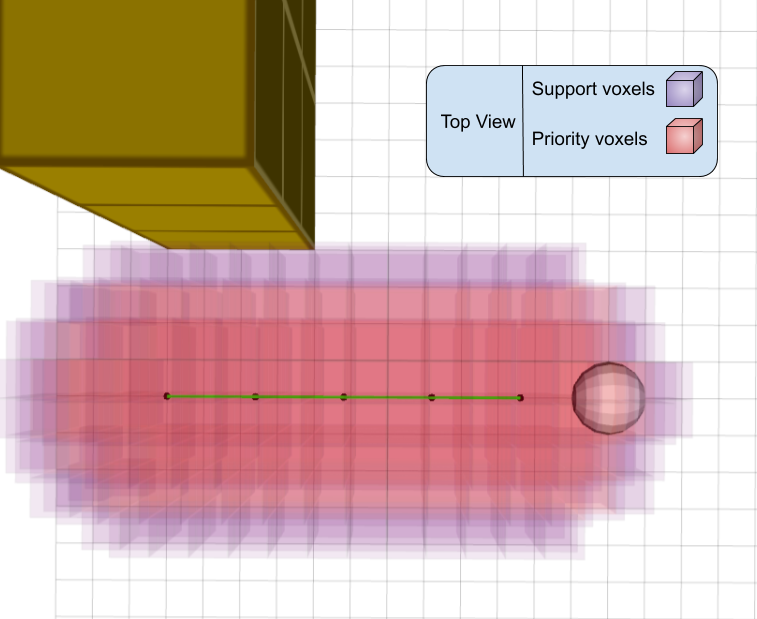}
\caption{Each tentacle has its own set of Support $S^v$ (magenta) and Priority $P^v$ (red) voxels inside the robot-centered grid. Tentacles are evaluated based on the occupancy in these voxels. If the occupied voxel is in $S^v$, its weight $\beta$ has higher value when it is closer to the tentacle. The weight gets its maximum when the voxel is in $P^v$.}
\label{fig:support_priority}
\end{figure}

In the robot-centered grid, the subset of voxels in the close range of each tentacle, are classified as either Support $S^{v}$ or Priority $P^{v}$, corresponding to "Support and Classification" areas in \cite{von2008driving}. These voxels are extracted as in the Eq. set (\ref{eqn:support}), based on the distance thresholds ${\tau^{S^{v}}}$ and ${\tau^{P^{v}}}$ where ${\tau^{S^{v}}} > {\tau^{P^{v}}}$. Each tentacle $j$ has its own set of Support and Priority voxels which are determined by the closest sampling point, $p_{min} \in T_j$ which satisfies the Eq. (\ref{eqn:pmin}). Hence, the set, $\Upsilon$, which contains support, $S^{v}$, and priority, $P^{v}$, voxels for all tentacles can be defined as $\Upsilon = \{S^{v}_j \cup P^{v}_j \quad | \quad j = 1,...,N^t\}$. The Fig. \ref{fig:support_priority} shows the extracted Priority and Support voxels for a particular tentacle.
\begin{subequations}
    \begin{align}
        v_i &\in
            \begin{cases}
                \label{eqn:pmin}
                P^{v} & if \quad |M^{-1}(o_i) - p_{min}| < \tau^{P^{v}} \\
                S^{v} & if \quad \tau^{P^{v}} < |M^{-1}(o_i) - p_{min}| < \tau^{S^{v}} 
            \end{cases}\\
        c_i &=
            \begin{cases}
                1 & if \quad v_i \in P^{v} \\
                0 & if \quad v_i \in S^{v}\text{,} \quad \text{where}
            \end{cases}
    \end{align}
    
    \begin{align}
        &S^{v} \cap P^{v} = \emptyset \quad \& \quad S^{v} \cup P^{v} \subseteq G, \\
        &|M^{-1}(o_i) - p_{min}| < |M^{-1}(o_i) - p_{k}|.
    \end{align}
    \label{eqn:support}
\end{subequations}

The occupancy weight for each voxel $\beta_i$ is calculated by the function in Eq. (\ref{eqn:weight}). For $\forall v_i \in P^{v}$ the equation gives the maximum weight $\beta_{max}$, since Priority voxels are the closest ones to the corresponding tentacle and any occupancy on them might imply a high-impact collision risk. When $v_i \in S^{v}$, the value of the weight become decreasing for farther voxels, where the rate can be adjusted by the parameter $\alpha_{\beta} > 0$.
\begin{equation} 
\begin{aligned}
\beta_i &= 
    \begin{cases} 
        \beta_{max} & if \quad v_i \in P^{v}\\
        \frac{\beta_{max}}{\alpha_{\beta} |M^{-1}(o_i) - p_{min}|} & if \quad v_i \in S^{v}.
        \label{eqn:weight}
    \end{cases}
\end{aligned}
\end{equation}~\\

\subsection{Tentacle Evaluation} \label{sec:tent_eval}
	
In every cycle of the algorithm, each tentacle $j$ is evaluated by five heuristic metrics derived from the path planning literature. In this paper, we address these metrics as Navigability $\Pi^{nav}_j$, Clearance $\Pi^{clear}_j$, Nearby Clutter $\Pi^{clut}_j$, Goal Closeness $\Pi^{close}_j$ and Smoothness $\Pi^{smo}_j$. Our interpretation of these heuristic functions are given in the following subsections: ~\\
	
\subsubsection{Navigability} \label{sec:navigability}

\begin{figure}[!ht]
\centering
\includegraphics[width=3.5in]{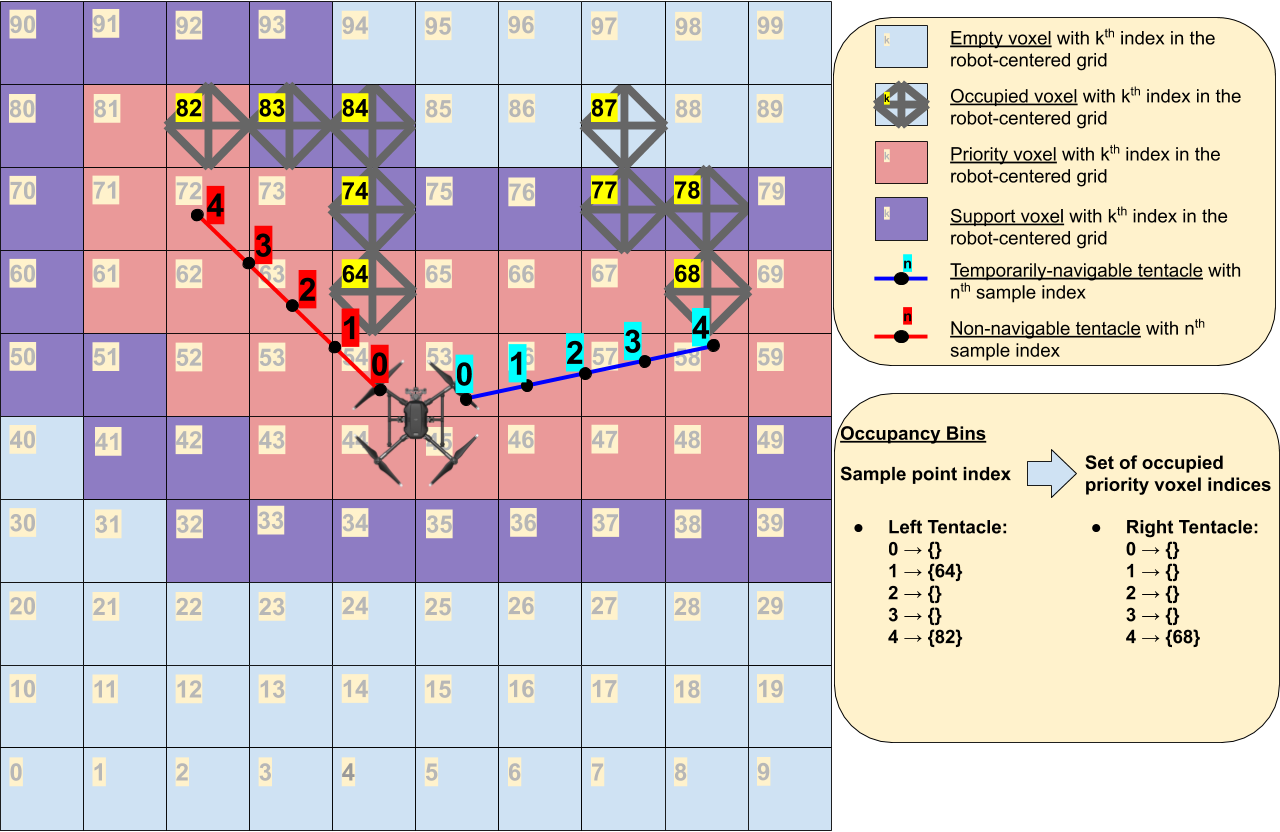}
\caption{Given the planar navigation scenario where the Support (magenta) and Priority (red) voxels are extracted for the two tentacles. Suppose that the crash distance is up to the second sampling point, the left tentacle becomes non-navigable since index of the sampling point, whose occupancy bin is not empty, is less than the crash distance. On the other hand, the right tentacle is classified as temporarily navigable due to the index of its first occupied bin is higher.}
\label{fig:navigability}
\end{figure}

For each tentacle $j$, $\Pi^{nav}_j$ assigns whether it is navigable ($1$), non-navigable ($0$) or temporarily navigable ($-1$) using the Eq. (\ref{eqn:navigability}). Here, the variable $l^t$ is the tentacle length. The crash distance threshold $\tau^{crash}$ can be adjusted by the rate parameter $\alpha_{crash} > 0$ as in Eq. (\ref{eqn:lobs1}). $l^{obs}_j$ in Eq. (\ref{eqn:lobs2}) is the distance from the first sampling point to the first occupied sampling point at $k^{obs}$ which satisfies the Eq. (\ref{eqn:lobs3}) given the occupancy error threshold $\tau^{D_{err}}$. The function $H_{k_j}$ projects the occupancy information of the Priority voxels onto the sampling points on the corresponding tentacle. To do that, first, the occupied Priority voxels, corresponding to the sampling point $k$ on the tentacle $j$, are determined. This is equivalent to find $v_{i_j}$'s in the Eq. (\ref{eqn:projection1}) and form the occupancy bins as in the example shown in Fig. \ref{fig:navigability}. Then, the projection function, $H_{k_j}$, is computed by the Eq. (\ref{eqn:projection1}) where the constraints are given in the Eq. (\ref{eqn:projection2}) and (\ref{eqn:projection3}).
\begin{subequations} 
    \begin{align}
        \label{eqn:navigability}
        \Pi^{nav}_j &= 
        \begin{cases} 
            1, \hspace{0.1em} & if \quad l^{obs}_j = l^t_j\\
            0, \hspace{0.1em} & if \quad l^{obs}_j < \tau^{crash} \\
            -1, \hspace{0.1em} & if \quad \tau^{crash} < l^{obs}_j < l^t_j
	    \end{cases}
	\end{align}
    where
    \begin{align}
        \label{eqn:lobs1}
        \tau^{crash} &= \frac{l^t_j}{\alpha^{crash}}\\
        \label{eqn:lobs2}
        l^{obs}_j &= \frac{l^t_j k^{obs}_j}{n^s}\\ 
        \label{eqn:lobs3}
        k^{obs}_j &= \min_j k_j\text{,} \quad \text{s.t.} \quad H_{k_j} > \tau^{D_{err}} \quad \forall k_j\\
	    \label{eqn:projection1}
        H_{k_j} &= \sum_{v_{i_j}}^{} 1\\
        \label{eqn:projection2}
        v_{i_j} &= (o_{i_j}, \beta_{i_j}, s_{i_j}, c_{i_j}) \in P^{v}\\
        \label{eqn:projection3}
        M^{-1}(o_{i_j}) &= p_{min} \in T_j, \quad s.t. \quad A(o_{i_j}) > 0.
	\end{align}
\end{subequations}

\subsubsection{Clearance} \label{sec:clearance}
$\Pi^{clear}_j$ reflects proximity of an obstacle on the tentacle. It is obtained by the ratio of $l^{obs}_j$ and the tentacle length $l^t_j$ as shown in Eq. (\ref{eqn:close}). The value range of the function changes from 0 (totally clear path) to 1 (occupied) based on the closest occupancy determined by the variable $l^{obs}_j$ which is already calculated while obtaining Navigability function.
\begin{equation} 
    \label{eqn:close}
	\begin{aligned}
	    \Pi^{clear}_j = 1 - \frac{l^{obs}_j}{l^t_j}.
	\end{aligned}
\end{equation}

\subsubsection{Nearby Clutter} \label{sec:clutter}
In order to evaluate the nearby clutter value $\Pi^{clut}_j$ for each tentacle $j$, the total weight $\Omega^{tot}_j$ and the total occupancy weight $\Omega^{obs}$ of Priority and Support voxels are calculated as in Eq. (\ref{eqn:clutter}).
\begin{subequations} 
	\begin{align}
	    \Pi^{clut}_j &= \frac{\Omega^{obs}}{\Omega^{tot}}\text{,} \quad \text{where,} \\
	    \Omega^{tot} &= \sum_{v_i}^{} \beta_i\\
	    \Omega^{obs} &= \sum_{v_i}^{} \beta_i A(o_i)\\
	    v_i &= (o_i, \beta_i, s_i, c_i) \in P^{v} \cup S^{v}.
    \end{align}
    \label{eqn:clutter}
\end{subequations}

\subsubsection{Goal Closeness} \label{sec:closeness} $\Pi^{close}_j$ is calculated by the Euclidean distance between a specified sampling point on the tentacle $p^{R}_s$ and the goal point $p^{goal}$ such that:
\begin{equation} 
	\begin{aligned}
	    \Pi^{close}_j = |p^{W}_s - p^{goal}|, \quad p^{R}_s \in T_j.
	\end{aligned}
\end{equation}
	
\subsubsection{Smoothness} \label{sec:smoothess}
$\Pi^{smo}_j$ is used for smoother tentacle transitions. The function assigns lower values to the tentacles which are closer to the previously selected tentacle $j = {best}$ as shown in the following equation. Similar as in the Goal Closeness function, $p^{R}_{k_j}$ is the specified sampling point on the tentacle $j$.
\begin{equation} 
	\begin{aligned}
	    \Pi^{smo}_j = |p^{R}_{k_j} - p^{R}_{k_{best}}|, \quad p^{R}_j \in T_j.
	\end{aligned}
	\label{eqn:smooth}
\end{equation}

\subsection{Tentacle Selection and Execution} \label{sec:tent_exe}

The cost function of each tentacle, $F_j$, is calculated by the weighted sum of four heuristic functions, $\Pi^{clear}_j, \Pi^{clut}_j, \Pi^{close}_j,\Pi^{smo}_j$ and the adjusted weights $\lambda^{clear}, \lambda^{clut}, \lambda^{close}, \lambda^{smo}$ respectively shown in the Eq. (\ref{eqn:value_func}). The tentacle $j$ which is evaluated as the minimum of $F_j$ and classified as completely or temporary navigable by $\Pi^{nav}_j$ is selected as the best tentacle as in the Eq. \ref{eqn:best_tentacle}.
\begin{equation}
    \label{eqn:value_func}
	\begin{aligned}
	    F_j = \lambda^{clear} \Pi^{clear}_j + \lambda^{clut} \Pi^{clut}_j \\
	+ \lambda^{close} \Pi^{close}_j + \lambda^{smo} \Pi^{smo}_j
	\end{aligned}
\end{equation}

\begin{equation}
    \label{eqn:best_tentacle}
	\begin{aligned}
	    j^{best} = arg\min_j F_j, \quad \forall j.
	\end{aligned}
\end{equation}

To determine the next robot position, we consider kinematic constraints of the robot such as maximum lateral and angular speeds. Instead of sending the first sampling point on the selected tentacle to the motion controller, we interpolate the point between current robot position and the sampling point at the crash distance of the selected tentacle. At each processing time $d_t$, the computed pose command is sent to the motion control unit where the lower level actuation is executed.

\subsection{Implementation Details} \label{sec:off_on_param}

We implement the proposed algorithm and the data structures in ROS Kinetic using C++. The pseudo-code is demonstrated in Algorithm \ref{algo} which enables the autonomous navigation in an unknown map perceived by the robot's sensors. In this context, we assume that the global positioning and odometry information of the robot and the goal(s) are available throughout the navigation. ~\\
\begin{algorithm}[ht!]
    \SetKwData{Left}{left}
    \SetKwData{Begin}{begin}
    \SetKwData{Up}{up}
    \SetKwFunction{Union}{Union}
    \SetKwFunction{Begin}{begin}
    \SetKwInOut{Input}{input}
    \SetKwInOut{Output}{output}
    \SetKwProg{Main}{main()}{}{end}
    \SetKwProg{Begin}{begin}{}{end}
    \SetKwFunction{FRecurs}{FnRecursive}%

    \Input{global coordinate frame $W$, goal point $p^{goal}$, point cloud data $D$, robot parameters $\chi^{R}$, ofline parameters $\chi^{off}$, online parameters $\chi^{on}$}
    \Begin{}
    {
        $A \leftarrow$ InitializeLinearGrid($D$, $\chi^{R}$, $\chi^{off}$); \\
        $Q \leftarrow$ GenerateTentacles($\chi^{R}$, $\chi^{off}$); \\
    	$\Upsilon \leftarrow$ ExtractSupportPriorityVoxels($\chi^{off}$, $Q$); \\
    	\While{$goalNotReached$ or $t < T_{limit}$}
    	{
    		$A \leftarrow$ UpdateLinearGrid($D$, $\chi^{R}$, $\chi^{off}$); \\
    				
    		\For{each tentacle $j$}
    		{
    		    $H_j, \Omega^{tot}_j, \Omega^{obs}_j \leftarrow$ UpdateOccInfo($\chi^{off}$, $\chi^{on}$, $\Upsilon$, $A$); \\
    		    $\Pi^{nav}_j \leftarrow$ UpdateNavigability($\chi^{off}$, $\chi^{on}$, $H_j$); \\
    			$\Pi^{clear}_j \leftarrow$ UpdateClearance($\chi^{off}$, $\Pi^{nav}_j$); \\
    			$\Pi^{clut}_j \leftarrow$ UpdateClutter($\Omega^{tot}_j$, $\Omega^{obs}_j$); \\
    			$\Pi^{close}_j \leftarrow$ UpdateCloseness(W, $\chi^{R}$, $\chi^{on}$, $p^{goal}$); \\
    			$\Pi^{smo}_j \leftarrow$ UpdateSmoothness($\chi^{R}$, $j^{best}$); \\

    			$F_{j} \leftarrow$ UpdateCost($\Pi^{nav}_j$, $\Pi^{clear}_j$, $\Pi^{clut}_j$, $\Pi^{close}_j$, $\Pi^{smo}_j$); \\
    		}
    		$j^{best} \leftarrow$ SelectBestTentacle($F_j$); \\
    		$\chi^{R} \leftarrow$ ExecuteMotion($\chi^{R}$, $j^{best}$); \\
    	}
    }
    \caption{Tentacle-based reactive navigation}
    \label{algo}
\end{algorithm}

Given as the input to the framework, structure of robot parameters $\chi^R$ includes volumetric, and kinematic information of the robot along with occupancy sensor specifications as described in Table \ref{table:params}. In order to enable utilization across robotic platforms, instead of considering exact volume of the robot, we adopt a bounding box model. The maximum lateral and angular velocity parameters affect the tentacle generation process. Similarly, the resolution and the range information of the navigation sensor define the size of the robot-centered 3D grid. ~\\
\begin{table}[!ht]
	\caption{Parameters}
	\begin{center}\label{table:params}
	    \begin{tabular}{|p{3cm} | p{5cm}|}
	        \hline
			{\bf Robot Parameters $\chi^{R}$} & {\bf Description}\\[0.5pt]
			\hline
            ${w^R, {l^R}, {h^R}}$ & Width, length, height of the robot \\
            \hline
            ${v_{lat}}$ & Max forward lateral velocity of the robot \\
            \hline
            ${\omega_{\varphi}, \omega_{\theta}, \omega_{\psi}}$ & Max angular velocity of the robot in yaw, pitch and roll \\
            \hline
            $d^s$ & Resolution of the navigation sensor \\
            \hline
            $\rho_x, \rho_y, \rho_z$ & Maximum range of the navigation sensor in $x$, $y$ and $z$ axes. \\
			\hline
			\hline
			{\bf Offline Parameters $\chi^{off}$} & \\[0.5pt]
			\hline
			${d^v}$ & Voxel dimension\\
            \hline
            $n^{v}_{\{x,y,z\}}$ & Number of grid voxels in each axes\\
            \hline
            $n_{\varphi}, n_{\theta}, n_{\psi}$ & Number of tentacles in yaw-pitch-roll\\
            \hline
            $n^s$ & Number of sampling points on a tentacle\\
            \hline
            $l^t$ & Tentacle length\\
            \hline
            $\varphi, \theta, \psi$ & Covered angle of tentacles in yaw, pitch and roll\\
            \hline
            ${\tau^{P}}, {\tau^{S}}$ & Distance thresholds with respect to Priority and Support voxels\\
			\hline
			$\beta_{max}$ & Max occupancy weight of Priority voxels\\
			\hline
			$\alpha_{\beta}$ & Occupancy weight scale of Priority and Support voxels\\
			\hline
			\hline
			{\bf Online Parameters $\chi^{on}$} & \\[0.5pt]
			\hline
			$\alpha^{crash}$ & Crash distance \\
            \hline
            $\lambda^{clear}$ & Clearance weight \\
            \hline
            $\lambda^{clut}$ & Nearby clutter weight \\
            \hline
            $\lambda^{close}$ & Goal closeness weight\\
            \hline
            $\lambda^{smo}$ & Smoothness weight\\
            \hline
			\end{tabular}
	\end{center}
\end{table}

The remaining input parameters, which directly affect the performance of the proposed navigation algorithm, are grouped into two categories and named as offline $\chi^{off}$ and online $\chi^{on}$ as given in Table \ref{table:params}. Since the reactive nature of the algorithm is empowered by the fast computation, the offline parameters are adjusted only before the navigation. On the other hand, online parameters can be updated during the navigation without causing much computational burden but to improve the performance. In essence, the general form of the tentacles and the robot-centered grid are formed by $\chi^{off}$ while navigation preferences such as greediness towards the goal or timidness while avoiding obstacles are tuned by $\chi^{on}$.~\\

Before the main navigation loop, the algorithm begins with the initialization of the robot-centered grid structure, which consist of a two linear array of size $(N^v)$. First array allocates memory for the occupancy information projected into the grid. The second one keeps positions of the voxel centers to enable mapping between 3D and linear indices. Then, the tentacles are generated by defining sampling points and their group (such as linear, circular, etc.) structure. The whole sampling points, $(x,y,z)$, are stored in a 2D vector of size $(N^tn^s)$, where each group of sampling points corresponds to the same tentacle. Having the robot-centered grid and tentacle information, Support and Priority voxels $v=(o, \beta, s, c)$ are extracted and kept in a 2D array of size $(N^tn^{SP})$ where $n^{SP} \subseteq N^v$. Hence, the order of growth of the whole pre-navigation functions can be given as $O(N^tn^{v})$. ~\\

The reactive navigation algorithm iterates until all of the goal points are reached or the time limit is exceeded. In the first step of each iteration, the linear occupancy grid is updated with the most recent point cloud information, $D$, which contains $N^D$ data point. This takes $O(N^D)$ processing time in our implementation. Then for each tentacle $j$ where $j \in \{1,...,N^t\}$, the heuristic functions are calculated. Since these functions are evaluated based on the sampling points, each of these functions also have a loop of size equal to the number of samples $n^s$. Therefore, computation time of all cost functions is bounded by $O(N^tn^s)$ where the best tentacle selection is $O(N^t)$. In the last step of each iteration, the execution of the pose command is performed by the controller developed by \cite{lee2010geometric} to generate the rotor actuation of the UAV in the physics-based simulations.

\section{Results} \label{sec:results}

\begin{figure}[!ht]
    \centering
    \includegraphics[width=3.5in]{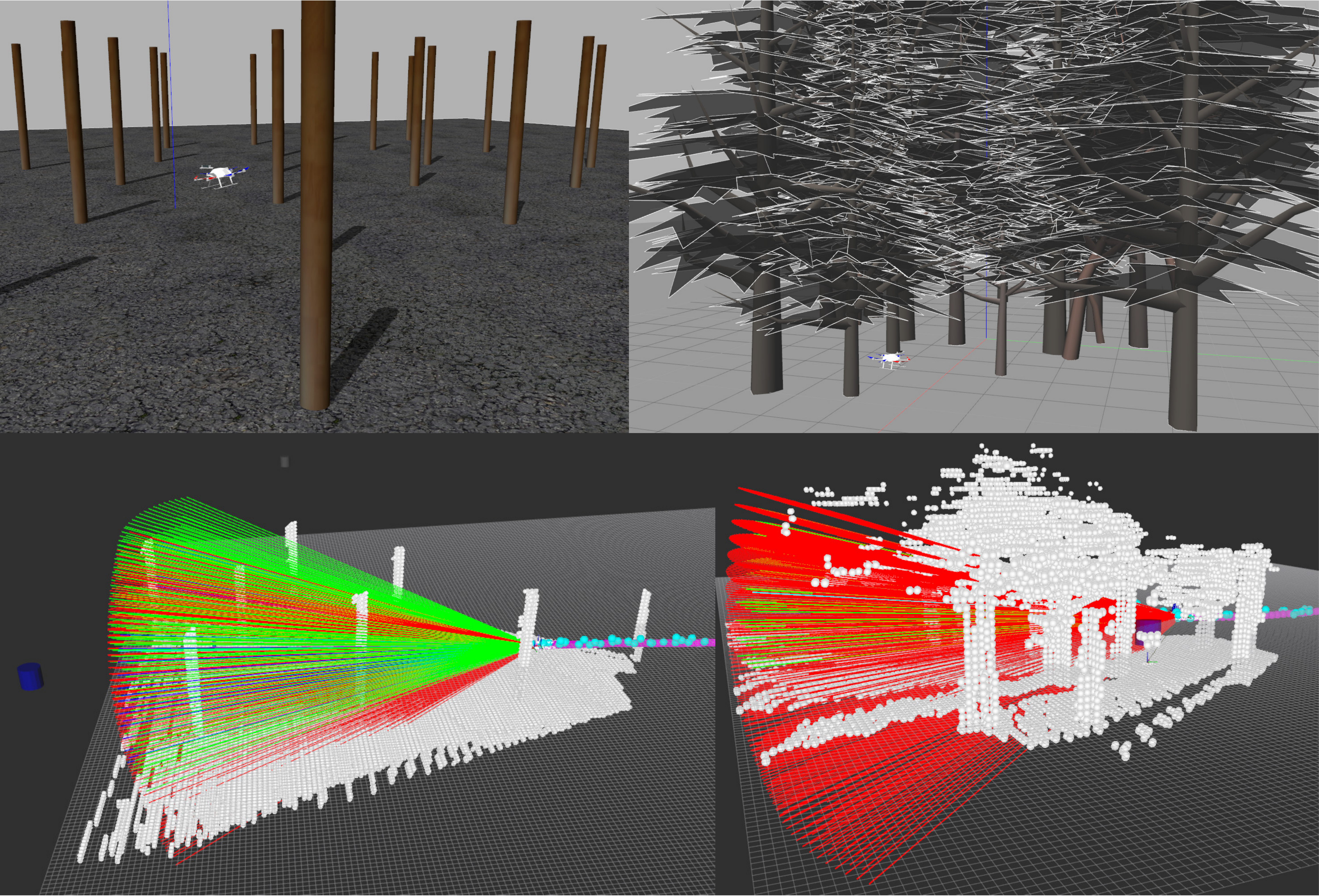}
    \caption{For the benchmark, two types of maps in Gazebo environment are used. (Top left) The first type  consists of cylindrical obstacles in $20x20m^2$ area. (Top right) The second type of map contains tree shaped obstacles whose density is $0.2trees/{m^2}$ inside of a $10x10m^2$ area. (Bottom) Rviz is used to observe status of the navigation including the occupancy, the trajectory of the robot, formation and navigability of tentacles.}
    \label{fig:simulation}
\end{figure}

 For the benchmark, two types of maps in Gazebo environment, which are available in the code repository of \cite{usenko2017real}, are used. The first type, shown in the top left of the Fig. \ref{fig:simulation}, consists of cylindrical obstacles in $20x20m^2$ area. We keep the same goal positions, as in \cite{usenko2017real}, which are determined to maximize the travelled distance. The second type of map, provided by the "$forest\_gen$" ROS package \cite{oleynikova2016continuous}, contains tree shaped obstacles whose density is $0.2trees/{m^2}$ inside of a $10x10m^2$ area. To enable rotor dynamics in our simulations, the AscTec Firefly model is used from the "$rotors\_simulator$" package \cite{furrer2016rotors} where the RGB-D sensor is mounted on the robot. ~\\

Before running the simulations, the $\chi^{R}, \chi^{on}, \chi^{off}$ are adjusted based on the robot model, sensor specifications and the navigation task. The range of the sensor regulates the tentacles' length and covered angles along yaw and pitch. Hence, $t^l, \varphi, \theta$ are set to $10m$, $60^o$ and $45^o$ respectively. The priority distance threshold $\tau^{P}$ is adjusted to $0.4m$ to encircle the bounding box of the robot while the support distance is set approximately twice more, $\tau^{P} = 1m$, empirically. Having specified the tentacle length and priority distance, the number of sampling points on a tentacle is assigned to {\bf $n^s=30$} in order to keep the robot inside of the priority voxels throughout the tentacle. The occupancy weight scale of the priority and support voxels is adjusted to $\alpha_{\beta}=10$. The max occupancy weight is set to $\beta_{max}=1$ to keep the occupancy weights in the range of $[0,1]$. ~\\

To analyze the effect of the remaining offline parameters on the computation time, 3 sets of simulations are performed and the results are given in the Table \ref{table:analysis}. Having the sensor with the resolution of $0.15m$, we test the voxel dimension $d^v$ for $0.2m$ and $0.1m$. In order to match the grid dimensions with the tentacles' length, the number of voxels in each axis $n^v_{x,y,z}$ are doubled when the $d^v$ is scaled down to half. This increases the total number of voxels in the grid by 8 times. Reflectively, the computation time of the initialization process of the linear grid, when $d^v=0.1$, is measured 8 times more than when $d^v=0.2$. The second and third column of the Table \ref{table:analysis} demonstrate the linear relationship between the number of tentacles and the total computation time of the "GenerateTentacles" and "ExtractSupportPriorityVoxels" steps. As expected, the computation time is measured twice as much when the $N^t$ is doubled. The duration of the main iteration steps, especially for the "UpdateOccInfo" and "UpdateHeuristics" steps, are harder to analyze since the calculations also depends on the momentary environment around the robot. Nevertheless, the statistical computation times, shown in Table \ref{table:analysis}, indicate logical results with respect to the changes in $d^v$ and $n^v_{x,y,z}$. Moreover, the total processing time of each simulation set proves that the algorithm is capable of running within the range of frequency from $10$ to $60$ $Hz$ successfully. ~\\
\begin{table}[!ht]
	\caption{Average computation time statistics of the initialization and the main iteration steps of the algorithm with respect to the voxel dimension $d^v$ and number of tentacles $N^t$}
	\begin{center}\label{table:analysis}
	    \begin{tabular}{|p{3.2cm}|p{1.3cm}|p{1.3cm}|p{1.4cm}|}
	        \hline
			{\bf Initialization Steps} & {\bf Time [s]} & {\bf Time [s]} & {\bf Time [s]} \\[0.5pt]
			 & {\bf $d^v=0.2$} {\bf $N^t=651$} & {\bf $d^v=0.1$} {\bf $N^t=651$} & {\bf $d^v=0.1$} {\bf $N^t=1271$} \\
			\hline
            InitializeLinearGrid & $0.2$ & $0.11$ & $0.11$ \\
            \hline
            GenerateTentacles + & $2.8$ & $24.03$ & $46.83$ \\
            ExtractSupportPriorityVoxels & & & \\
            \hline
            \hline
            {\bf Total} & $2.82$ & $24.14$ & $46.95$ \\
            \hline
            \hline
            {\bf Main Iteration Steps} & {\bf Time [ms]} & {\bf Time [ms]} & {\bf Time [ms]} \\[0.5pt]
            \hline
            UpdateOccInfo & $6.77$ & $11.51$ & $9.46$ \\
            \hline
            UpdateHeuristics & $7.95$ & $79.33$ & $108.07$ \\
            \hline
            SelectBestTentacle & $0.002$ & $0.002$ & $0.003$ \\
            \hline
            ExecuteMotion & $0.02$ & $0.01$ & $0.02$ \\
            \hline
            \hline
            {\bf Total} & $14.73$ & $91$ & $117.55$ \\
            \hline
			\end{tabular}
	\end{center}
\end{table}

Our proposed algorithm is also benchmarked with the implementation of the work in \cite{usenko2017real} within the 10 maps (cylinders map + 9 forest maps). For the first comparison, we keep their default parameters as they provided in their code repository. For the second one, we increase only the number of optimization points, $C$, from 7 to 9 since it gives the best result according to their paper \cite{usenko2017real}. To test the robustness, all configurations are run 10 times for each map without changing any parameter. ~\\

As discussed earlier in this section, the offline parameters can be mostly adjusted by the robot and the occupancy sensor specifications. For the benchmark simulations, we set the offline parameters same as given and keep them fix for  all maps. On the other hand, the tuning of the online parameters highly depends on the given task and the environment due to the reactive nature of our proposed algorithm. Hence, based on the goal location relative to the obstacles and the density of the each map, the online parameters are manually tuned to get the best performance. Although this might be considered as a weakness of the algorithm, the overall tuning process becomes quite straightforward when the logic behind the heuristic functions are comprehended. Considering that and as a future work, the online tuning process can be learned from the previous experiences and automatically tuned during the navigation. ~\\

\begin{figure}[!ht]
\centering
\includegraphics[width=3.5in]{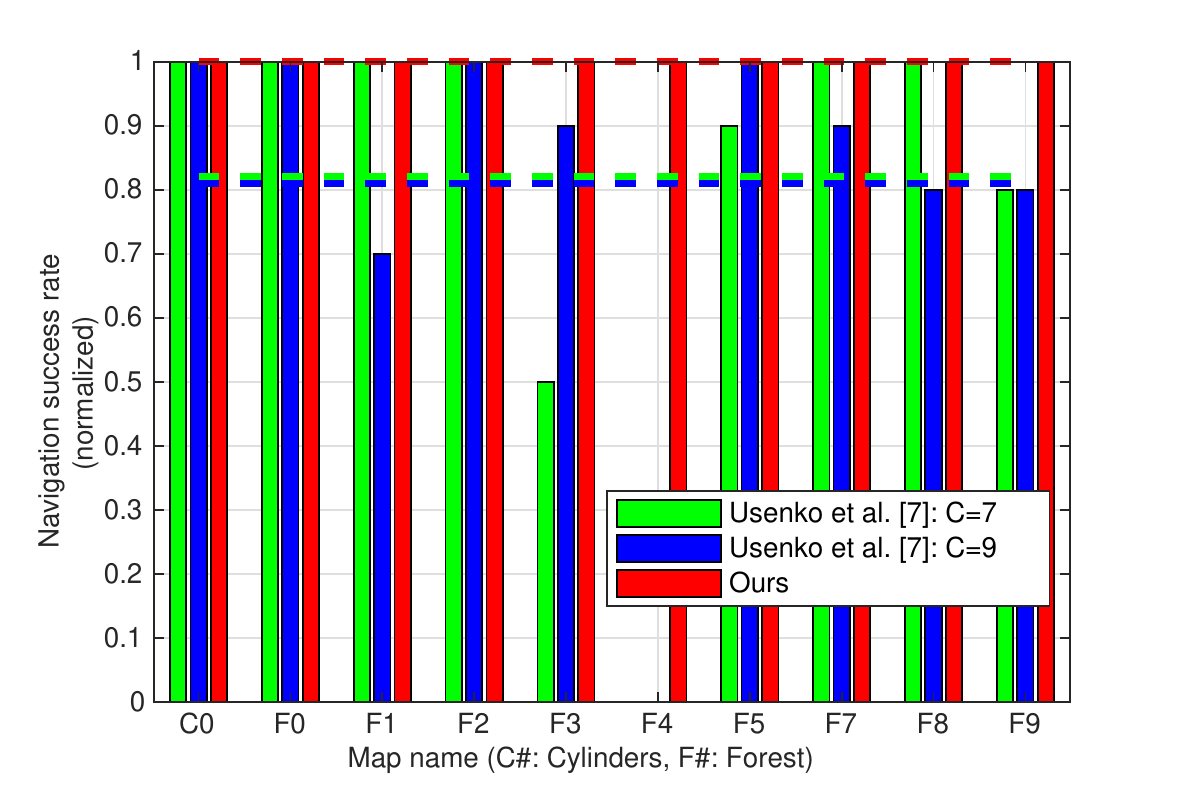}
\includegraphics[width=3.5in]{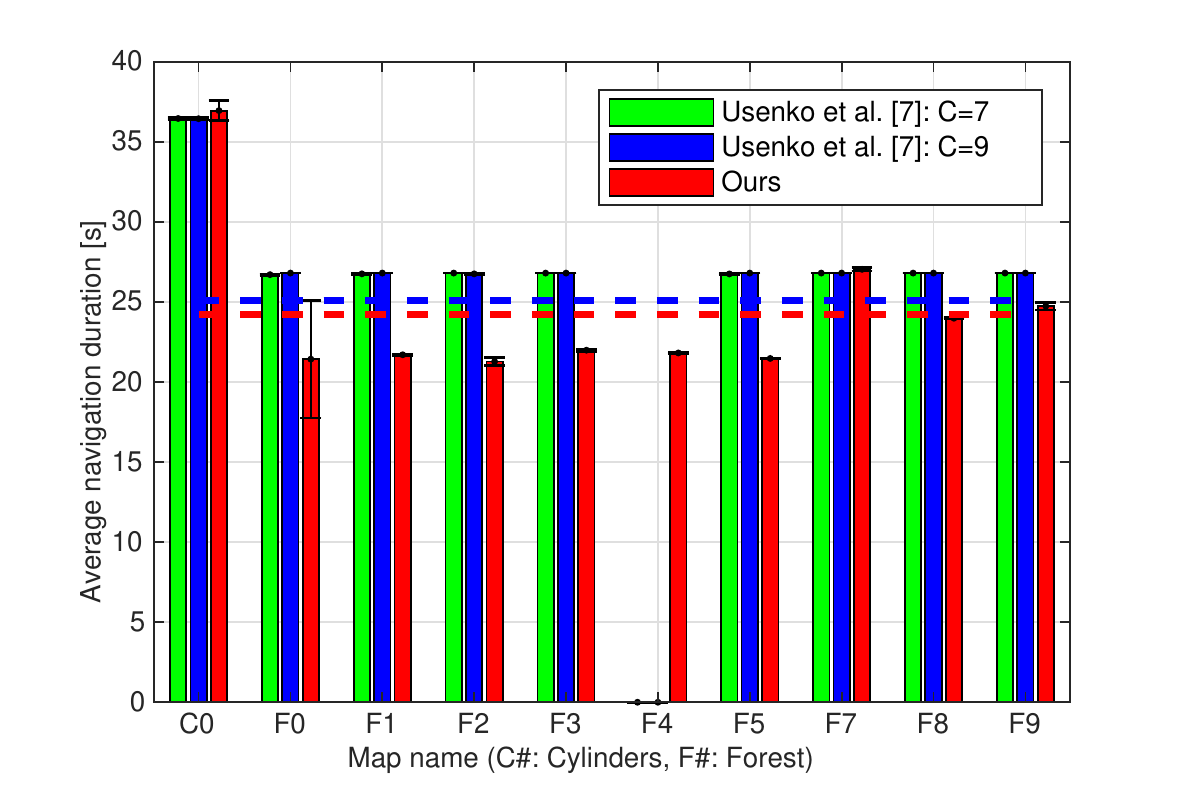}
\includegraphics[width=3.5in]{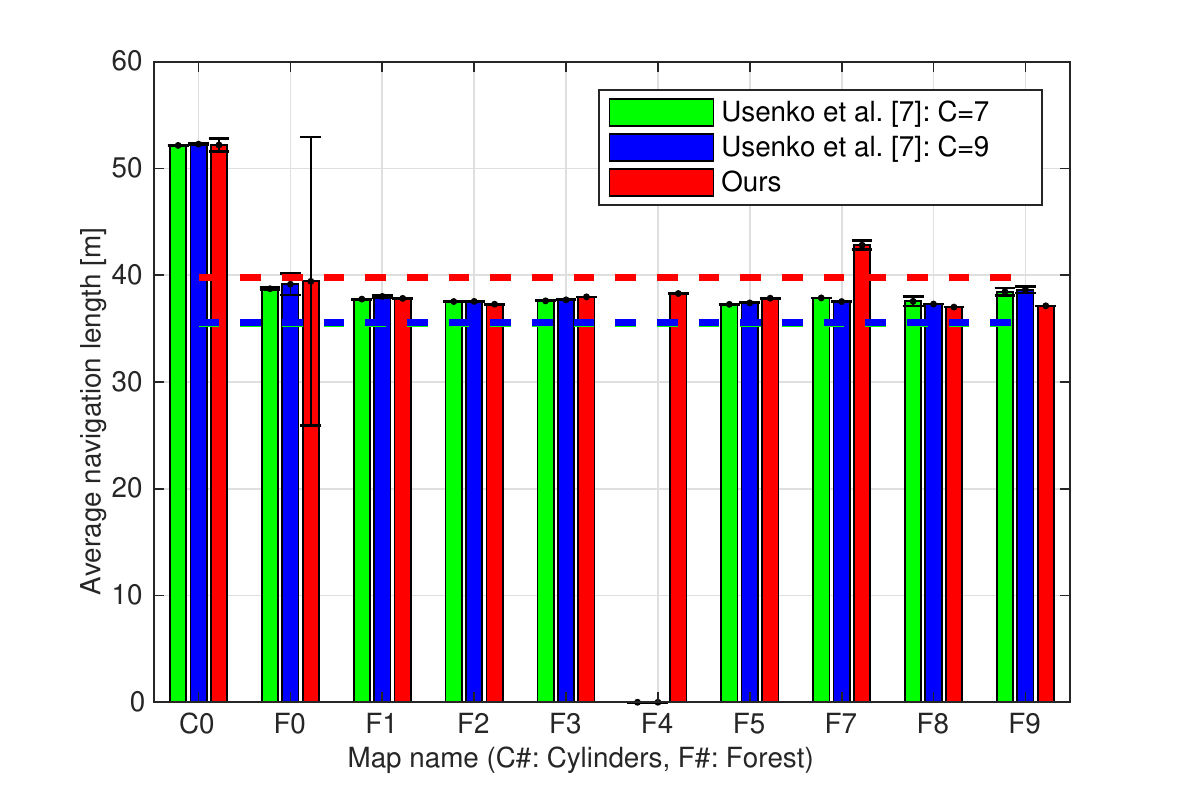}
\caption{Based on (top) navigation success rate, (middle) duration and (bottom) the path length, the statistical performances of the proposed algorithm in 10 different maps are benchmarked with the implementation of the work in \cite{usenko2017real}. The dashed lines shows the average values for all maps.}
\label{fig:benchmark}
\end{figure}

Overall, the simulation results reveal that our proposed algorithm has higher success rate and enables faster navigation as demonstrated in the first two plots in Fig. \ref{fig:benchmark}. Remarkably, our method succeeds in all successive trials for all maps, while both of the configurations of the state-of-art algorithm are failed at all in the Forest4 map. Although our average path length is slightly higher than the other algorithm, the third plot shows that ours is capable of finding shorter paths for the half of the maps. Noting that our algorithm is reactive and does not keep the global map or path history, its instability of finding the optimal path is quite expected. Besides, we adjust the online parameters to prioritize the safety over greediness to the goal.

\section{Conclusion}
In this paper, we present a reactive navigation algorithm for 3D environments that does not rely on a global map information. This is achieved by the pre-determined group of points, named as tentacles, which sample the space around the robot. The robot-centered grid structure is formed to keep the occupancy information. In order to evaluate the tentacles and select the best possible next point, five heuristic functions are defined. This paper also introduces offline and online parameters to enhance the reactive navigation performance. The approach of tuning these parameters are explained along with the other implementation details including computational complexity analysis. We perform the physics-based simulations using benchmark datasets. Overall, our proposed reactive algorithm outperforms two configurations of the state-of-art method in terms of success rate and navigation duration.

\section*{Acknowledgment}
This research was supported by the Department of Homeland Security as part of the National Infrastructure Protection Plan (NIPP) Security and Resilience Challenge and the Northeastern University's Global Resilience Institute seed funding program. This research is also supported by the  National  Science  Foundation  under  Award  No.  1544895, 1928654, 1935337, 1944453.



\bibliographystyle{IEEEtran}
\bibliography{refs}

\begin{thebibliography}{10}
\providecommand{\url}[1]{#1}
\csname url@samestyle\endcsname
\providecommand{\newblock}{\relax}
\providecommand{\bibinfo}[2]{#2}
\providecommand{\BIBentrySTDinterwordspacing}{\spaceskip=0pt\relax}
\providecommand{\BIBentryALTinterwordstretchfactor}{1}
\providecommand{\BIBentryALTinterwordspacing}{\spaceskip=\fontdimen2\font plus
\BIBentryALTinterwordstretchfactor\fontdimen3\font minus
  \fontdimen4\font\relax}
\providecommand{\BIBforeignlanguage}[2]{{%
\expandafter\ifx\csname l@#1\endcsname\relax
\typeout{** WARNING: IEEEtran.bst: No hyphenation pattern has been}%
\typeout{** loaded for the language `#1'. Using the pattern for}%
\typeout{** the default language instead.}%
\else
\language=\csname l@#1\endcsname
\fi
#2}}
\providecommand{\BIBdecl}{\relax}
\BIBdecl

\bibitem{lin2018autonomous}
Y.~Lin, F.~Gao, T.~Qin, W.~Gao, T.~Liu, W.~Wu, Z.~Yang, and S.~Shen,
  ``Autonomous aerial navigation using monocular visual-inertial fusion,''
  \emph{Journal of Field Robotics}, vol.~35, no.~1, pp. 23--51, 2018.

\bibitem{oleynikova2016continuous}
H.~Oleynikova, M.~Burri, Z.~Taylor, J.~Nieto, R.~Siegwart, and E.~Galceran,
  ``Continuous-time trajectory optimization for online {UAV} replanning,'' in
  \emph{IEEE/RSJ International Conference on Intelligent Robots and Systems
  (IROS)}, 2016.

\bibitem{beul2018fast}
M.~Beul, D.~Droeschel, M.~Nieuwenhuisen, J.~Quenzel, S.~Houben, and S.~Behnke,
  ``Fast autonomous flight in warehouses for inventory applications,''
  \emph{IEEE Robotics and Automation Letters}, vol.~3, no.~4, pp. 3121--3128,
  2018.

\bibitem{escobar2018r}
H.~D. Escobar-Alvarez, N.~Johnson, T.~Hebble, K.~Klingebiel, S.~A. Quintero,
  J.~Regenstein, and N.~A. Browning, ``R-{ADVANCE}: Rapid adaptive prediction
  for vision-based autonomous navigation, control, and evasion,'' \emph{Journal
  of Field Robotics}, vol.~35, no.~1, pp. 91--100, 2018.

\bibitem{oleynikova2019open}
H.~Oleynikova, C.~Lanegger, Z.~Taylor, M.~Pantic, A.~Millane, R.~Siegwart, and
  J.~Nieto, ``An open-source system for vision-based micro-aerial vehicle
  mapping, planning, and flight in cluttered environments,'' \emph{arXiv
  preprint arXiv:1812.03892}, 2019.

\bibitem{gao2018online}
F.~Gao, W.~Wu, Y.~Lin, and S.~Shen, ``Online safe trajectory generation for
  quadrotors using fast marching method and {B}ernstein basis polynomial,'' in
  \emph{2018 IEEE International Conference on Robotics and Automation
  (ICRA)}.\hskip 1em plus 0.5em minus 0.4em\relax IEEE, 2018, pp. 344--351.

\bibitem{usenko2017real}
V.~Usenko, L.~von Stumberg, A.~Pangercic, and D.~Cremers, ``Real-time
  trajectory replanning for {MAV}s using uniform {B}-splines and a 3{D}
  circular buffer,'' in \emph{2017 IEEE/RSJ International Conference on
  Intelligent Robots and Systems (IROS)}.\hskip 1em plus 0.5em minus
  0.4em\relax IEEE, 2017, pp. 215--222.

\bibitem{campos2019autonomous}
L.~Campos-Mac{\'\i}as, R.~Aldana-L{\'o}pez, R.~de~la Guardia, J.~I.
  Parra-Vilchis, and D.~G{\'o}mez-Guti{\'e}rrez, ``Autonomous navigation of
  {MAV}s in unknown cluttered environments,'' \emph{arXiv preprint
  arXiv:1906.08839}, 2019.

\bibitem{mohta2018fast}
K.~Mohta \emph{et~al.}, ``Fast, autonomous flight in {GPS}-denied and cluttered
  environments,'' \emph{Journal of Field Robotics}, vol.~35, no.~1, pp.
  101--120, 2018.

\bibitem{karaman2011sampling}
S.~Karaman and E.~Frazzoli, ``Sampling-based algorithms for optimal motion
  planning,'' \emph{The International Journal of Robotics Research}, vol.~30,
  no.~7, pp. 846--894, 2011.

\bibitem{hart1968formal}
P.~E. Hart, N.~J. Nilsson, and B.~Raphael, ``A formal basis for the heuristic
  determination of minimum cost paths,'' \emph{IEEE Transactions on Systems
  Science and Cybernetics}, vol.~4, no.~2, pp. 100--107, 1968.

\bibitem{von2008driving}
F.~Von~Hundelshausen, M.~Himmelsbach, F.~Hecker, A.~Mueller, and H.-J.
  Wuensche, ``Driving with tentacles: Integral structures for sensing and
  motion,'' \emph{Journal of Field Robotics}, vol.~25, no.~9, pp. 640--673,
  2008.

\bibitem{himmelsbach2009team}
M.~Himmelsbach, F.~Von~Hundelshausen, T.~Luettel, M.~Manz, A.~Mueller,
  S.~Schneider, and H.~Wuensche, ``Team {M}u{CAR}-3 at {C}-{ELROB} 2009,'' in
  \emph{Proceedings of 1st workshop on field robotics, civilian European land
  robot trial}, 2009.

\bibitem{himmelsbach2011autonomous}
M.~Himmelsbach, T.~Luettel, F.~Hecker, F.~von Hundelshausen, and H.-J.
  Wuensche, ``Autonomous off-road navigation for {M}u{CAR}-3,''
  \emph{KI-K{\"u}nstliche Intelligenz}, vol.~25, no.~2, pp. 145--149, 2011.

\bibitem{cherubini2012new}
A.~Cherubini, F.~Spindler, and F.~Chaumette, ``A new tentacles-based technique
  for avoiding obstacles during visual navigation,'' in \emph{Robotics and
  Automation (ICRA), 2012 IEEE International Conference on}.\hskip 1em plus
  0.5em minus 0.4em\relax IEEE, 2012, pp. 4850--4855.

\bibitem{cherubini2014autonomous}
A.~Cherubini, F.~Spindler, and F.~Chaumette, ``Autonomous visual navigation and
  laser-based moving obstacle avoidance,'' \emph{IEEE Transactions on
  Intelligent Transportation Systems}, vol.~15, no.~5, pp. 2101--2110, 2014.

\bibitem{alia2015local}
C.~Alia, T.~Gilles, T.~Reine, and C.~Ali, ``Local trajectory planning and
  tracking of autonomous vehicles, using clothoid tentacles method,'' in
  \emph{Intelligent Vehicles Symposium (IV), 2015 IEEE}.\hskip 1em plus 0.5em
  minus 0.4em\relax IEEE, 2015, pp. 674--679.

\bibitem{mouhagir2016markov}
H.~Mouhagir, R.~Talj, V.~Cherfaoui, F.~Guillemard, and F.~Aioun, ``A {M}arkov
  decision process-based approach for trajectory planning with clothoid
  tentacles,'' in \emph{IEEE Intelligent Vehicles Symposium (IV 2016)}, 2016,
  pp. 1254--1259.

\bibitem{mouhagir2017trajectory}
H.~Mouhagir, V.~Cherfaoui, R.~Talj, F.~Aioun, and F.~Guillemard, ``Trajectory
  planning for autonomous vehicle in uncertain environment using evidential
  grid,'' \emph{IFAC-PapersOnLine}, vol.~50, no.~1, pp. 12\,545--12\,550, 2017.

\bibitem{zhang2017formation}
M.~Zhang, ``Formation flight and collision avoidance for multiple {UAV}s based
  on modified tentacle algorithm in unstructured environments,'' \emph{PloS
  ONE}, vol.~12, no.~8, p. e0182006, 2017.

\bibitem{khelloufi2017tentacle}
A.~Khelloufi, N.~Achour, R.~Passama, and A.~Cherubini, ``Tentacle-based moving
  obstacle avoidance for omnidirectional robots with visibility constraints,''
  in \emph{Intelligent Robots and Systems (IROS), 2017 IEEE/RSJ International
  Conference on}.\hskip 1em plus 0.5em minus 0.4em\relax IEEE, 2017, pp.
  1331--1336.

\bibitem{hornung2013octomap}
A.~Hornung, K.~M. Wurm, M.~Bennewitz, C.~Stachniss, and W.~Burgard,
  ``Octo{M}ap: An efficient probabilistic 3{D} mapping framework based on
  octrees,'' \emph{Autonomous Robots}, vol.~34, no.~3, pp. 189--206, 2013.

\bibitem{lee2010geometric}
T.~Lee, M.~Leok, and N.~H. McClamroch, ``Geometric tracking control of a
  quadrotor {UAV} on {SE} (3),'' in \emph{49th IEEE Conference on Decision and
  Control (CDC)}.\hskip 1em plus 0.5em minus 0.4em\relax IEEE, 2010, pp.
  5420--5425.

\bibitem{furrer2016rotors}
F.~Furrer, M.~Burri, M.~Achtelik, and R.~Siegwart, ``Rotor{S}—{A} modular
  {G}azebo {MAV} simulator framework,'' in \emph{Robot Operating System
  (ROS)}.\hskip 1em plus 0.5em minus 0.4em\relax Springer, 2016, pp. 595--625.

\end{thebibliography}

\end{document}